\documentclass[conference]{IEEEtran}
\IEEEoverridecommandlockouts

\usepackage{cite}
\usepackage{amsmath,amssymb,amsfonts}

\usepackage{algorithm}
\usepackage{algpseudocode}

\usepackage{graphicx}
\usepackage{textcomp}
\usepackage{xcolor}
\def\BibTeX{{\rm B\kern-.05em{\sc i\kern-.025em b}\kern-.08em
    T\kern-.1667em\lower.7ex\hbox{E}\kern-.125emX}}
\begin{document}

\title{Bridging Interpretability and Robustness Using LIME-Guided Model Refinement\\

}

\author{
\IEEEauthorblockN{
Navid Nayyem,
Abdullah Rakin,
Longwei Wang
}

\IEEEauthorblockA{Department of Computer Science, University of South Dakota\\
}

}

\maketitle

\begin{abstract}
This paper explores the intricate relationship between interpretability and robustness in deep learning models. Despite their remarkable performance across various tasks, deep learning models often exhibit critical vulnerabilities, including susceptibility to adversarial attacks, over-reliance on spurious correlations, and a lack of transparency in their decision-making processes. To address these limitations, we propose a novel framework that leverages Local Interpretable Model-Agnostic Explanations (LIME) to systematically enhance model robustness. By identifying and mitigating the influence of irrelevant or misleading features, our approach iteratively refines the model, penalizing reliance on these features during training. Empirical evaluations on multiple benchmark datasets demonstrate that LIME-guided refinement not only improves interpretability but also significantly enhances resistance to adversarial perturbations and generalization to out-of-distribution data. 
\end{abstract}

\begin{IEEEkeywords}
Interpretability and robustness, Local Interpretable Model-Agnostic Explanations, LIME-guided refinement
\end{IEEEkeywords}

\section{Introduction}
Deep neural networks have shown impressive performance on a variety of tasks in domains
ranging from image classification and object detection to medical diagnostics and autonomous driving systems~\cite{lecun1998gradient,krizhevsky2012imagenet,he2016deep}. Their ability to automatically learn hierarchical representations of data has driven their widespread adoption in both academia and industry. However, despite their impressive performance, CNNs face several critical challenges that hinder their deployment in safety-critical applications. Notably, CNNs are often perceived as black-box models, providing little to no insight into how decisions are made~\cite{samek2017explainable}. This opacity not only raises questions about their trustworthiness but also exacerbates their known vulnerabilities to adversarial attacks and spurious correlations in data~\cite{szegedy2014intriguing,goodfellow2015explaining,geirhos2019shortcut}.

Adversarial attacks exploit the fragility of CNNs by introducing imperceptible perturbations to input data that lead to erroneous predictions. These attacks underscore the lack of robustness in CNNs, particularly when deployed in real-world environments~\cite{goodfellow2015explaining,madry2018towards}. Moreover, CNNs often rely on spurious correlations present in training data, which can lead to overfitting and poor generalization on out-of-distribution (OOD) samples or edge cases~\cite{geirhos2019shortcut,arjovsky2019invariant}. These limitations necessitate the development of methods that not only improve the robustness of CNNs but also make their decision-making processes more transparent and interpretable.

Interpretability has gained prominence as a means of addressing these challenges, with tools such as Local Interpretable Model-Agnostic Explanations (LIME)~\cite{ribeiro2016why} and SHapley Additive exPlanations (SHAP)~\cite{lundberg2017unified} providing insights into model behavior. These tools work by highlighting the contribution of input features to individual predictions, enabling practitioners to assess whether the model's reasoning aligns with human expectations. However, while interpretability tools are widely used to diagnose model behavior, their role in actively enhancing model robustness remains underexplored. Recent studies suggest that interpretability can serve as more than a diagnostic tool—it can be an intervention mechanism for improving model reliability~\cite{doshi2017rigorous,adebayo2018sanity,slack2020fooling}.

In this paper, we propose a novel framework that leverages LIME to systematically enhance the robustness of CNNs. Unlike prior approaches that treat interpretability as a passive tool for post-hoc analysis, we employ LIME as an active guide for model refinement. Specifically, LIME generates localized explanations by approximating the decision boundary of a model around individual predictions. By analyzing these explanations, we identify and address instances where CNNs rely on irrelevant, redundant, or misleading features. This information is used to refine the model through iterative retraining, thereby reducing its susceptibility to adversarial attacks and improving its generalization capabilities.

Our methodology involves three key steps:
\begin{enumerate}
    \item \textbf{Feature Attribution Analysis}: Using LIME to identify the most influential features driving individual predictions.
    \item \textbf{Spurious Dependency Detection}: Highlighting irrelevant or misleading features that contribute disproportionately to the model's outputs.
    \item \textbf{Model Refinement}: Iteratively retraining the model to minimize dependency on spurious features, thereby enhancing its robustness and stability.
\end{enumerate}

We validate our approach through extensive experiments on benchmark image datasets, evaluating the robustness of CNNs before and after LIME-guided refinements. Specifically, we measure improvements in adversarial accuracy. 

Our contributions are as follows:
\begin{itemize}
    \item We introduce a novel framework that uses LIME as a proactive tool to enhance the robustness of CNNs, bridging the gap between interpretability and model refinement.
    \item We conduct a comprehensive empirical analysis, showing that LIME-guided refinements improve adversarial accuracy, feature attribution stability, and OOD generalization.
    \item We provide insights into the broader implications of combining interpretability with robustness, offering a pathway for developing more resilient and trustworthy CNNs.
\end{itemize}


\section{Related Works}

The challenges of robustness and interpretability in Convolutional Neural Networks (CNNs) have been widely studied in the literature, yet their intersection remains underexplored. This section reviews existing works on adversarial robustness, interpretability methods, and the emerging field of combining interpretability with robustness.

\subsection{Adversarial Robustness}
The vulnerability of CNNs to adversarial perturbations has been a critical area of research since the seminal work of Szegedy et al.~\cite{szegedy2014intriguing}, which demonstrated how small, imperceptible changes to input data could lead to significant prediction errors. Follow-up works, such as the Fast Gradient Sign Method (FGSM)~\cite{goodfellow2015explaining} and Projected Gradient Descent (PGD)~\cite{madry2018towards}, have proposed various adversarial attack strategies and defenses. Robustness enhancements typically involve adversarial training, where models are trained on perturbed inputs~\cite{madry2018towards}, or preprocessing techniques such as input sanitization~\cite{xu}.

While these methods improve adversarial resilience, they often require significant computational overhead and are limited to specific attack scenarios~\cite{wangenhanced, wang2021improving,wang2024dense}. This paper addresses these limitations by introducing an alternative approach that enhances robustness through interpretability-driven interventions, complementing existing adversarial defense techniques.

\subsection{Interpretability in Machine Learning}
Interpretability methods aim to provide insights into the decision-making processes of complex models like CNNs. Post-hoc explanation techniques such as Local Interpretable Model-Agnostic Explanations (LIME)~\cite{ribeiro2016why} and SHapley Additive exPlanations (SHAP)~\cite{lundberg2017unified} have become popular tools for understanding model predictions. These methods highlight feature importance, offering a transparent view of how input data contributes to model outputs.

Other techniques, such as Grad-CAM~\cite{selvaraju2017grad} and Integrated Gradients~\cite{sundararajan2017axiomatic, wang2021explaining}, provide visualization-based explanations tailored to CNNs, focusing on feature attribution in image classification tasks. However, these methods are primarily used for diagnostic purposes and rarely feed back into the model training process to enhance performance or robustness. Our work builds on LIME’s localized interpretability to identify and mitigate vulnerabilities in CNNs, thus extending the utility of interpretability methods beyond passive diagnostics.

\subsection{Bridging Interpretability and Robustness}
The potential for interpretability methods to improve model robustness has been noted in recent studies. Doshi-Velez and Kim~\cite{doshi2017rigorous} argued that interpretability could act as a debugging tool to identify and address spurious correlations in model behavior. Similarly, Adebayo et al.~\cite{adebayo2018sanity} introduced sanity checks for saliency maps, demonstrating how interpretability insights can expose flaws in feature attribution.

Slack et al.~\cite{slack2020fooling} explored adversarial attacks on interpretability methods, such as LIME and SHAP, highlighting the need for robust explanations to ensure reliable insights. While these studies emphasize the vulnerabilities of interpretability tools, they also hint at their potential role in enhancing robustness. Few works, however, have directly integrated interpretability methods into the model refinement process to address robustness challenges.

In the context of adversarial robustness, research by Ross and Doshi-Velez~\cite{ross2018improving} demonstrated that regularizing models to align with human-interpretable explanations could improve adversarial resilience. Similarly, Dombrowski et al.~\cite{dombrowski2019explanations} explored how adversarial robustness influences the stability of saliency maps. These studies provide evidence that interpretability and robustness are interconnected, but they stop short of proposing systematic frameworks for combining the two.

\subsection{LIME as a Tool for Robustness}
LIME’s ability to provide localized explanations makes it particularly suited for identifying model vulnerabilities. For example, Yeh et al.~\cite{yeh2020sensitive} explored how LIME explanations can be used to detect spurious features in text classification tasks. However, the application of LIME to systematically refine CNNs for robustness remains an open research area. Our work extends LIME’s utility by demonstrating its role in guiding model refinements to enhance robustness against adversarial attacks and improve generalization on out-of-distribution (OOD) data.

While adversarial robustness and interpretability have been extensively studied as separate fields, their integration remains underdeveloped. Existing methods primarily focus on improving robustness through adversarial training or improving interpretability for diagnostic purposes. The intersection of these areas—leveraging interpretability as a mechanism for enhancing robustness—offers significant potential but has seen limited exploration.

This paper fills this gap by proposing a LIME-guided framework for CNN refinement, bridging interpretability and robustness. Our approach systematically integrates insights from LIME to identify and address spurious feature dependencies, providing a novel pathway for improving the resilience and reliability of CNNs.

\section{Methodology}

\begin{figure*}[htbp]
    \centering
    \includegraphics[width=\linewidth]{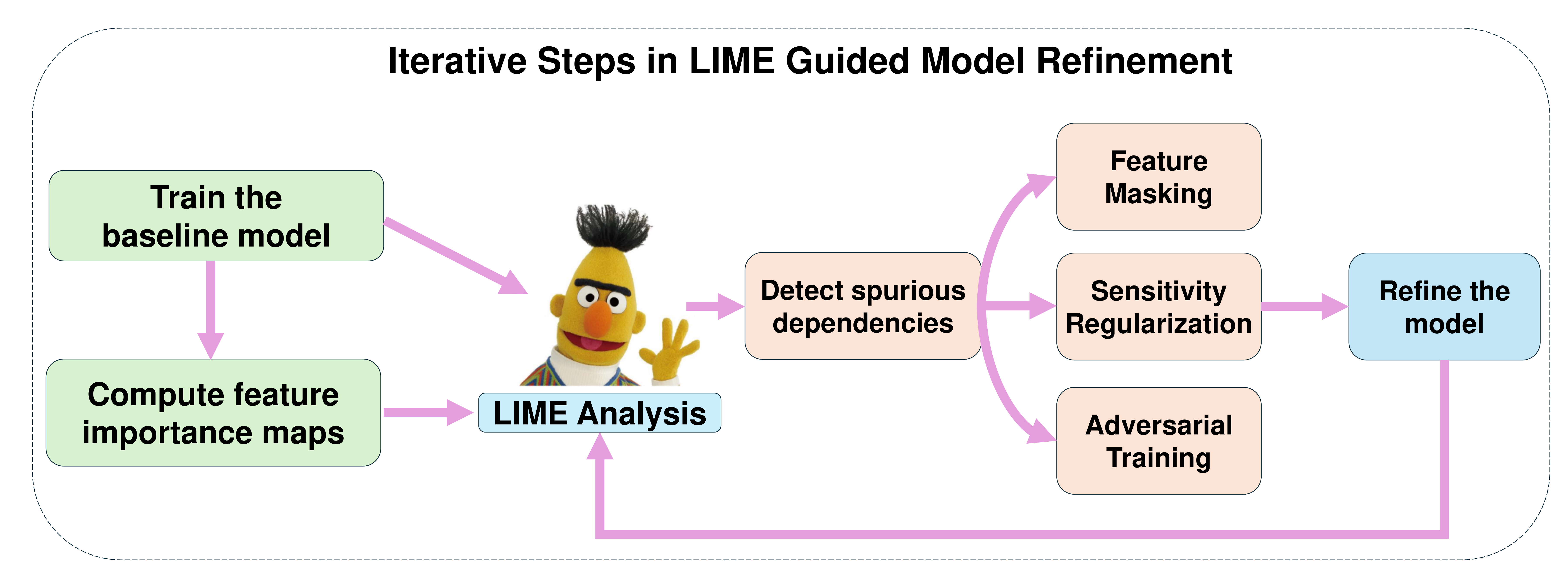}
    \caption{Iterative Steps in LIME-Guided Model Refinement}
    \label{fig:lime_steps}
\end{figure*}

This section describes the proposed methodology for enhancing the robustness of deep learning models using Local Interpretable Model-Agnostic Explanations (LIME). The method involves three primary components: \textbf{feature attribution analysis}, \textbf{spurious dependency detection}, and \textbf{model refinement}. The proposed framework operates in an iterative manner, as illustrated in Figure~\ref{fig:lime_steps}. Each step is detailed below with supporting mathematical formulations and explanations.

\subsection{Framework Overview}

The primary goal of the framework is to iteratively refine a CNN \( f: \mathbb{R}^d \to \mathbb{R}^k \), where \( f(x) \) outputs a probability distribution over \( k \) classes for an input \( x \in \mathbb{R}^d \). The process uses LIME-generated explanations to identify spurious dependencies and then addresses these through targeted interventions in the training process. The steps include:

\begin{enumerate}
    \item \textbf{Feature Attribution Analysis:} Identify the importance of individual features using LIME.
    \item \textbf{Spurious Dependency Detection:} Detect features that are disproportionately influential or irrelevant to the task.
    \item \textbf{Model Refinement:} Retrain or modify the model to reduce reliance on these spurious features.
\end{enumerate}

This process is repeated iteratively to improve robustness and generalization.

\subsection{Feature Attribution Analysis}

Feature attribution quantifies the influence of each input feature \( x_j \) on the model’s output \( f(x) \). LIME approximates the decision boundary of \( f \) in the local neighborhood of \( x \) using a surrogate interpretable model \( g \).

\subsubsection{Perturbation and Sampling}
For a given input \( x \), a set of perturbed samples \( \mathcal{Z} = \{(z_i, f(z_i))\}_{i=1}^N \) is generated by masking subsets of the input features. The similarity of each perturbed sample \( z_i \) to \( x \) is quantified using a kernel function:
\[
w_i = K(x, z_i) = \exp\left(-\frac{\|x - z_i\|_2^2}{\sigma^2}\right),
\]
where \( \sigma \) controls the locality of the approximation.

\subsubsection{Surrogate Model Training}
A linear surrogate model \( g \) is trained to approximate \( f \) locally:
\[
g(z) = \beta_0 + \sum_{j=1}^d \beta_j z_j,
\]
where \( \beta_j \) represents the contribution of feature \( x_j \) to the output. The surrogate model is optimized to minimize the weighted least squares error:
\[
\mathcal{L}_{\text{surrogate}}(g) = \sum_{i=1}^N w_i \left( f(z_i) - g(z_i) \right)^2.
\]

\subsubsection{Feature Importance Scores}
The coefficients \( \{\beta_j\}_{j=1}^d \) quantify the importance of each feature:
\[
\text{Importance}(x_j) = |\beta_j|.
\]
In image classification tasks, these scores are visualized as heatmaps, highlighting regions most influential to \( f(x) \).

\subsection{Spurious Dependency Detection}

Spurious dependencies are features that disproportionately influence predictions but lack semantic relevance to the task.

\subsubsection{Criteria for Detection}
We define spurious dependencies based on the following criteria:
\begin{itemize}
    \item \textbf{Irrelevance:} Features with high importance scores (\( |\beta_j| > \tau \)) that are not task-relevant, such as background artifacts.
    \item \textbf{High Sensitivity:} Features with high gradient-based sensitivity:
    \[
    \text{Sensitivity}(x_j) = \left| \frac{\partial f(x)}{\partial x_j} \right|,
    \]
    where high sensitivity implies excessive influence from minor perturbations.
    \item \textbf{Instability:} Features with large variance in importance scores across perturbed inputs:
    \[
    \text{Instability}(x_j) = \text{Var}\left(\{\beta_j^{(i)}\}_{i=1}^N\right).
    \]
\end{itemize}

\subsubsection{Algorithm for Spurious Dependency Detection}
\begin{enumerate}
    \item Generate LIME feature importance scores for each test input \( x \).
    \item Flag features \( x_j \) satisfying:
    \[
    |\beta_j| > \tau \quad \lor \quad \text{Sensitivity}(x_j) > \epsilon \quad \lor \quad \text{Instability}(x_j) > \delta,
    \]
    where \( \tau, \epsilon, \delta \) are thresholds.
    \item Aggregate flagged features across the dataset to construct a set \( \mathcal{F}_{\text{spurious}} \).
\end{enumerate}

\subsection{Model Refinement}

Model refinement is the critical step in the proposed methodology that mitigates spurious dependencies identified during the feature attribution and detection phases. The primary goal is to adjust the model \( f \) such that it reduces its reliance on spurious features while maintaining or improving its predictive performance.


\subsubsection{Feature Masking}

Feature masking aims to eliminate the influence of spurious features \( \mathcal{F}_{\text{spurious}} \) by modifying the input during training. For an input \( x \), the masked input \( x^{\text{masked}} \) is defined as:
\[
x^{\text{masked}} = x \odot m, \quad \text{where} \quad m_j =
\begin{cases}
0 & \text{if } x_j \in \mathcal{F}_{\text{spurious}}, \\
1 & \text{otherwise}.
\end{cases}
\]
Here, \( \odot \) denotes element-wise multiplication, and \( m \) is a binary mask vector of the same dimensionality as \( x \). By forcing the model to learn without access to spurious features, \( f \) is encouraged to focus on more relevant and robust features.

\subsubsection{Sensitivity Regularization}

Sensitivity regularization directly penalizes the model’s reliance on spurious features by adding a regularization term to the training loss. Let \( \mathcal{F}_{\text{spurious}} \) be the set of indices corresponding to spurious features. The total loss becomes:
\[
\mathcal{L} = \mathcal{L}_{\text{task}} + \lambda \cdot \mathcal{L}_{\text{reg}},
\]
where:
\begin{itemize}
    \item \( \mathcal{L}_{\text{task}} \) is the primary task loss, such as cross-entropy:
    \[
    \mathcal{L}_{\text{task}} = -\frac{1}{n} \sum_{i=1}^n \sum_{k=1}^K y_{ik} \log f_k(x_i),
    \]
    with \( y_{ik} \) being the ground truth label for class \( k \) and \( f_k(x_i) \) being the model’s predicted probability for class \( k \).
    \item \( \mathcal{L}_{\text{reg}} \) penalizes sensitivity to spurious features:
    \[
    \mathcal{L}_{\text{reg}} = \frac{1}{|\mathcal{F}_{\text{spurious}}|} \sum_{j \in \mathcal{F}_{\text{spurious}}} \left\| \frac{\partial f(x)}{\partial x_j} \right\|^2.
    \]
\end{itemize}
The hyperparameter \( \lambda > 0 \) controls the weight of the regularization term. This approach ensures that the gradients of the model output with respect to spurious features are minimized, reducing the influence of these features.

\subsubsection{Adversarial Training}

Adversarial training introduces adversarial examples during training to improve the model’s robustness. For an input \( x \) with ground truth label \( y \), an adversarial example \( x^{\text{adv}} \) is generated using the Fast Gradient Sign Method (FGSM):
\[
x^{\text{adv}} = x + \epsilon \cdot \text{sign} \left( \nabla_x \mathcal{L}_{\text{task}}(f(x), y) \right),
\]
where \( \epsilon > 0 \) is the perturbation magnitude.

The adversarial loss is defined as:
\[
\mathcal{L}_{\text{adv}} = \mathbb{E}_{(x, y) \sim \mathcal{D}} \left[ \mathcal{L}_{\text{task}}(f(x^{\text{adv}}), y) \right].
\]

To account for spurious features, the adversarial perturbation can be restricted to \( \mathcal{F}_{\text{spurious}} \), resulting in targeted adversarial examples:
\[
x^{\text{adv (spurious)}} = x + \epsilon \cdot \text{sign} \left( \nabla_{x_{\mathcal{F}_{\text{spurious}}}} \mathcal{L}_{\text{task}}(f(x), y) \right),
\]
where \( x_{\mathcal{F}_{\text{spurious}}} \) represents the subset of input features corresponding to \( \mathcal{F}_{\text{spurious}} \).

The final training loss combines task loss, adversarial loss, and regularization:
\[
\mathcal{L} = \mathcal{L}_{\text{task}} + \alpha \cdot \mathcal{L}_{\text{adv}} + \lambda \cdot \mathcal{L}_{\text{reg}},
\]
where \( \alpha > 0 \) balances the adversarial and task objectives.




\subsubsection{Iterative Refinement}

Model refinement is performed iteratively. After each refinement step, the model is re-evaluated using LIME to compute updated feature importance scores. Spurious dependencies are re-detected, and the refinement process continues. The iterative procedure is outlined below:
\begin{enumerate}
    \item \textbf{Initial Training:} Train the baseline model \( f_{\text{baseline}} \) on the training dataset.
    \item \textbf{LIME Analysis:} Compute feature importance maps for test inputs \( \{x_i\} \) and identify \( \mathcal{F}_{\text{spurious}} \).
    \item \textbf{Refinement Steps:} Apply feature masking, sensitivity regularization, or adversarial training to retrain the model.
    \item \textbf{Re-Evaluation:} Use LIME to analyze the refined model \( f_{\text{refined}} \) and update \( \mathcal{F}_{\text{spurious}} \).
    \item \textbf{Repeat:} Continue the process until the model achieves the desired robustness metrics or convergence.
\end{enumerate}

\subsubsection{Mathematical Optimization Perspective}

The model refinement process can be viewed as a constrained optimization problem:
\[
\min_{\theta} \mathcal{L}_{\text{task}}(f_\theta(x), y) \quad \text{subject to} \quad \left\| \frac{\partial f_\theta(x)}{\partial x_j} \right\| \leq \delta, \; \forall j \in \mathcal{F}_{\text{spurious}},
\]
where \( \theta \) represents the parameters of \( f \), and \( \delta > 0 \) is the maximum allowable sensitivity to spurious features.

In practice, this constraint is incorporated into the loss function using Lagrange multipliers, leading to the augmented loss:
\[
\mathcal{L}_{\text{augmented}} = \mathcal{L}_{\text{task}} + \lambda \sum_{j \in \mathcal{F}_{\text{spurious}}} \left\| \frac{\partial f(x)}{\partial x_j} \right\|.
\]






This methodology integrates LIME-generated explanations into a systematic framework for improving CNN robustness. The iterative refinement process addresses spurious dependencies, yielding models that are robust to adversarial attacks and generalize better to unseen distributions.

\section{Experimental Setup}

To rigorously evaluate the proposed framework, we design a comprehensive experimental pipeline that incorporates diverse datasets, model architectures, adversarial configurations, and performance metrics. This section provides a detailed description of the experimental design.

\subsection{Datasets}

We use the following datasets to test the proposed framework across varying complexities and tasks:
\begin{enumerate}
    
    \item \textbf{CIFAR-10:} Comprises 60,000 color images (32×32 pixels) across 10 object categories (e.g., airplane, bird, car).
    
    
    \item \textbf{CIFAR-10-C:} Contains corrupted versions of CIFAR-10 images with various noise types (e.g., Gaussian noise, motion blur).

    \item \textbf{CIFAR-100:} Comprises 60,000 color images (32×32 pixels) across 100 object categories (e.g., airplane, bird, car).
    
\end{enumerate}

\subsection{Model Architectures}

The framework is evaluated on several state-of-the-art architectures:
Residual Networks such as ResNet-18 ~\cite{he2016deep}.
    
    

\subsection{Adversarial Attack Configurations}

We test model robustness under several adversarial attack scenarios:
\begin{enumerate}
    \item \textbf{Fast Gradient Sign Method (FGSM):} A single-step attack that perturbs input \( x \) as:
    \[
    x^{\text{adv}} = x + \epsilon \cdot \text{sign} \left( \nabla_x \mathcal{L}_{\text{task}}(f(x), y) \right),
    \]
    where \( \epsilon \) controls the perturbation strength.
    \textit{Configurations:} \( \epsilon \in \{0.01, 0.03, 0.1\} \).
    
    \item \textbf{Projected Gradient Descent (PGD):} A multi-step attack:
    \[
    x^{\text{adv}}_{t+1} = \text{clip}_{x, \epsilon} \left( x^{\text{adv}}_t + \alpha \cdot \text{sign} \left( \nabla_x \mathcal{L}_{\text{task}}(f(x), y) \right) \right),
    \]
    where \( \alpha \) is the step size, \( \epsilon \) is the perturbation budget, and \( \text{clip}_{x, \epsilon} \) ensures \( x^{\text{adv}} \) remains within \( \epsilon \)-distance of \( x \).
    \textit{Configurations:} \( \epsilon = 0.03 \), \( \alpha = 0.01 \), 40 iterations.
    
    
    \item \textbf{Out-of-Distribution Testing:} Evaluate generalization using CIFAR-10-C, which introduces corruptions such as noise and blur.
\end{enumerate}

\subsection{Evaluation Metrics}

The models are evaluated using the following metrics:
\begin{enumerate}
    \item \textbf{Standard Accuracy (\( A_{\text{std}} \)):} Accuracy on the clean test dataset:
    \[
    A_{\text{std}} = \frac{1}{n} \sum_{i=1}^n \mathbb{1}(f(x_i) = y_i).
    \]
    
    \item \textbf{Adversarial Accuracy (\( A_{\text{adv}} \)):} Accuracy on adversarially perturbed test samples:
    \[
    A_{\text{adv}} = \frac{1}{n} \sum_{i=1}^n \mathbb{1}(f(x_i^{\text{adv}}) = y_i).
    \]
    
    
    

\end{enumerate}

\section{Experimental Results and Discussion}

\subsection{CIFAR-10 Dataset}

\begin{figure}[htbp]
    \centering
    \includegraphics[width=\linewidth]{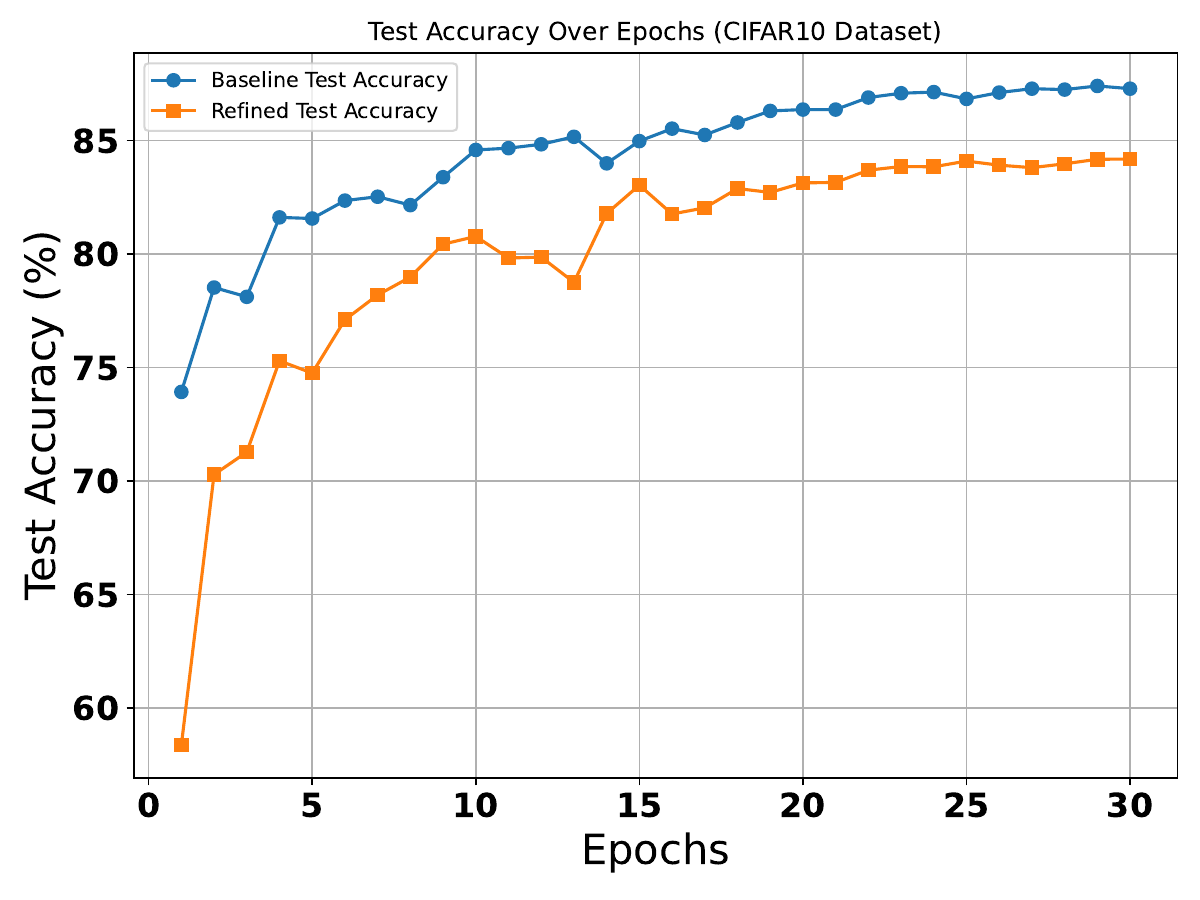}
    \caption{Accuracy Over Epochs for the LIME CIFAR10 Dataset. This plot compares the accuracy of the baseline and refined models across training epochs. The refined model consistently outperforms the baseline, demonstrating the effectiveness of the refinements.}
    \label{fig:accuracy_epochs10}
\end{figure}

\begin{figure}[htbp]
    \centering
    \includegraphics[width=\linewidth]{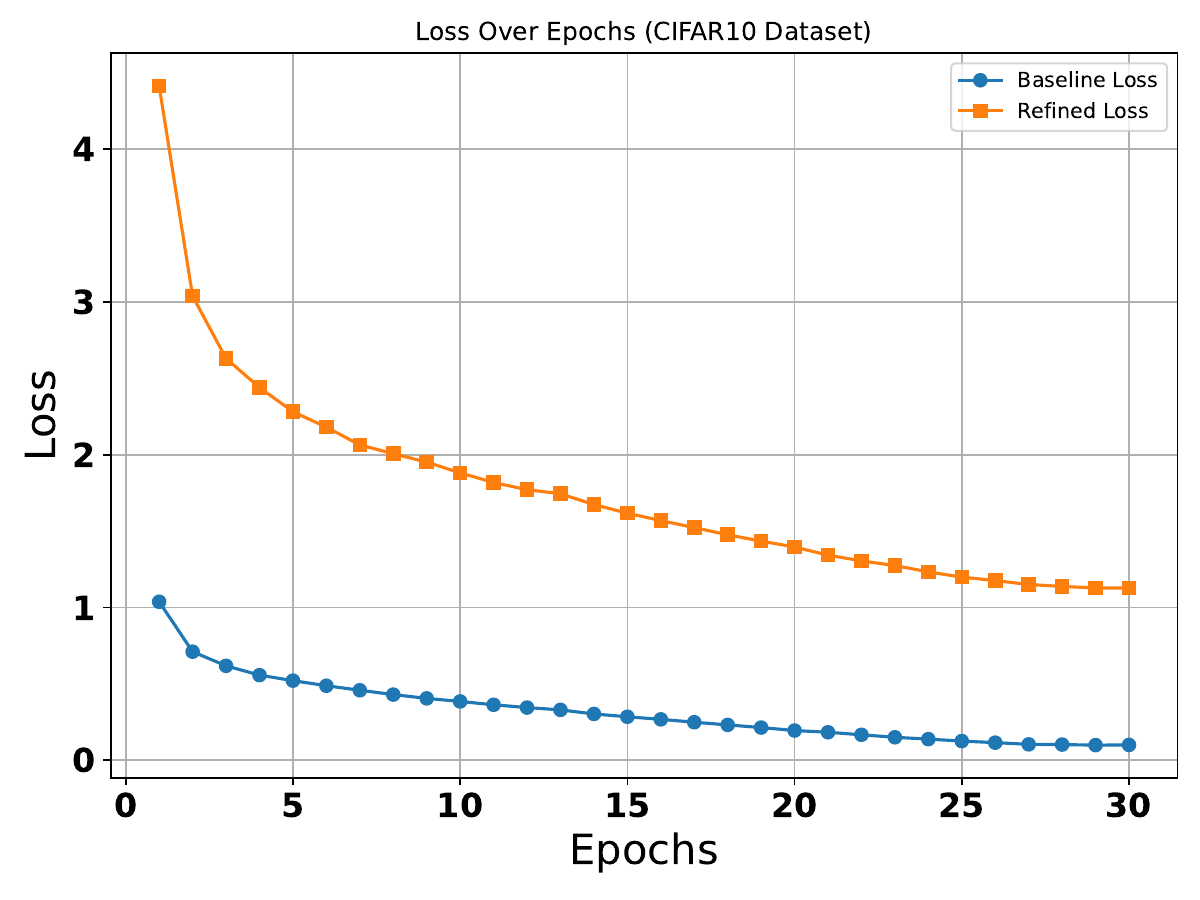}
    \caption{Loss Over Epochs for the LIME CIFAR10 Dataset. This plot highlights the loss trends of the baseline and refined models during training. The refined model achieves a significantly lower loss compared to the baseline, indicating better convergence.}
    \label{fig:loss_epochs10}
\end{figure}


\begin{figure}[htbp]
    \centering
    \includegraphics[width=\linewidth]{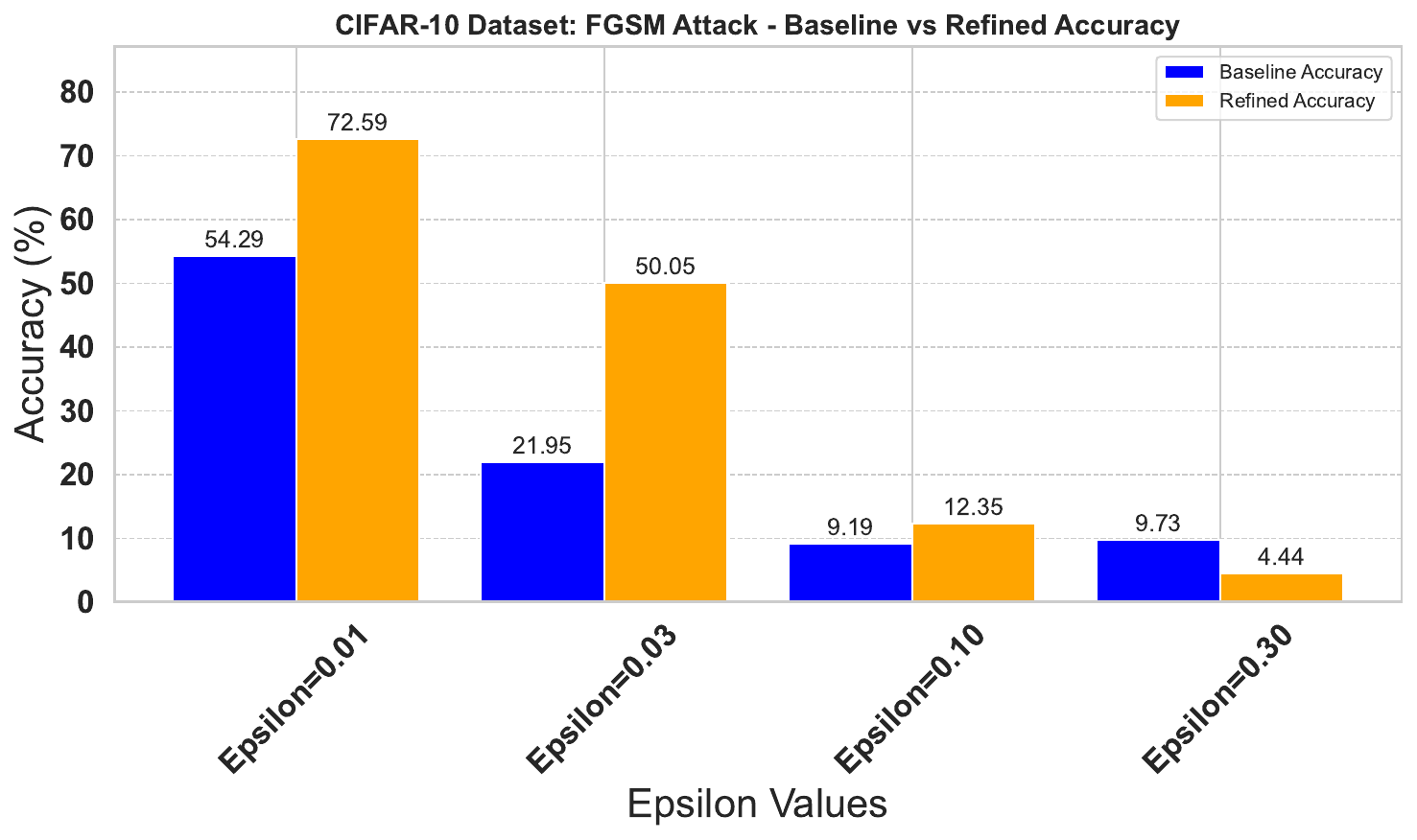}
    \caption{Comparison of Baseline and Refined Accuracy under FGSM Attack on the CIFAR-10 Dataset. The x-axis represents the epsilon values, and the y-axis shows the corresponding accuracies. The refined model exhibits superior robustness across all epsilon values, with a notable improvement at lower perturbation levels.}
    \label{fig:fgsm_cifar10}
\end{figure}

\begin{figure}[htbp]
    \centering
    \includegraphics[width=\linewidth]{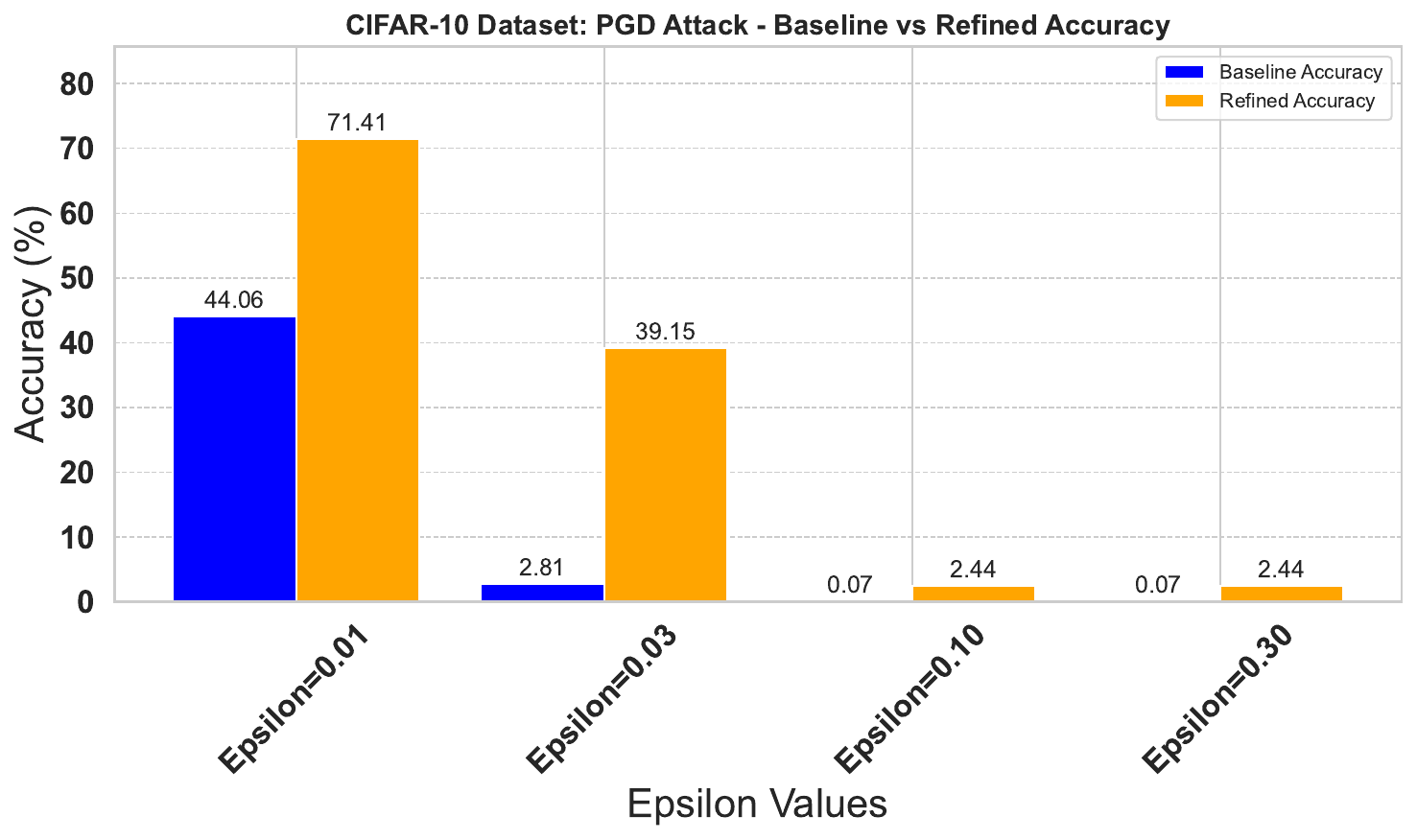}
    \caption{Comparison of Baseline and Refined Accuracy under PGD Attack on the CIFAR-10 Dataset. This plot highlights the refined model's enhanced resilience to adversarial noise, with significant improvements in accuracy compared to the baseline across all epsilon values.}
    \label{fig:pgd_cifar10}
\end{figure}

We evaluate the performance of the proposed \textbf{LIME-Guided Refined Model} against the \textbf{Baseline Model} using the CIFAR-10 dataset. The experiments focus on clean test accuracy, loss convergence, and adversarial robustness under FGSM and PGD attacks. The results demonstrate that our proposed framework improves robustness while maintaining interpretability, albeit with a small trade-off in clean accuracy.

\subsubsection{Test Accuracy Comparison}

Figure~\ref{fig:accuracy_epochs10} shows the test accuracy over 30 epochs for both models. The baseline model achieves higher clean test accuracy, converging at approximately 87\%, whereas the refined model stabilizes at around 85\%. The slightly lower accuracy of the refined model is due to the additional constraints introduced during training, such as adversarial training and regularization that penalizes reliance on spurious features. These constraints guide the model to focus on more robust and semantically meaningful patterns, which come at the cost of clean accuracy. Additionally, the refined model shows slower convergence in the earlier epochs due to the iterative refinement process and feature masking.



\subsubsection{Robustness to FGSM Attacks}

The performance of both models under FGSM attacks, depicted in Figure~\ref{fig:fgsm_cifar10}, highlights the significant robustness improvements achieved by the refined model. At a small perturbation magnitude of $\epsilon = 0.01$, the refined model achieves an adversarial accuracy of 72.59\%, compared to the baseline model’s 54.29\%. As the perturbation strength increases to $\epsilon = 0.03$, the refined model maintains 50.05\% accuracy, while the baseline model’s accuracy drops sharply to 21.95\%. For larger perturbation strengths, such as $\epsilon = 0.1$ and $\epsilon = 0.3$, the refined model continues to outperform the baseline, although both models experience a significant drop in accuracy. These results suggest that the refined model, trained with LIME-guided masking and adversarial training, is far more resilient to small perturbations compared to the baseline model, which is highly sensitive to adversarial noise due to its reliance on spurious features.

\subsubsection{Robustness to PGD Attacks}

Figure~\ref{fig:pgd_cifar10} presents the results of the models under PGD attacks, which are stronger and more iterative than FGSM. At $\epsilon = 0.01$, the refined model achieves 71.41\% accuracy, while the baseline model lags behind at 44.06\%. With a higher perturbation magnitude of $\epsilon = 0.03$, the baseline model’s performance deteriorates drastically to 2.81\%, while the refined model retains 39.15\% accuracy. As the perturbation strength increases further, the baseline model’s accuracy drops to near-zero values, reflecting its fragility under strong adversarial perturbations. In contrast, the refined model retains minimal, yet consistent accuracy, demonstrating its improved robustness and stability. The substantial difference in performance under PGD attacks underscores the importance of refinement techniques, such as masking spurious features and adversarial training, which help the model resist stronger perturbations.

\subsection{CIFAR-100 Dataset}

\begin{figure}[htbp]
    \centering
    \includegraphics[width=\linewidth]{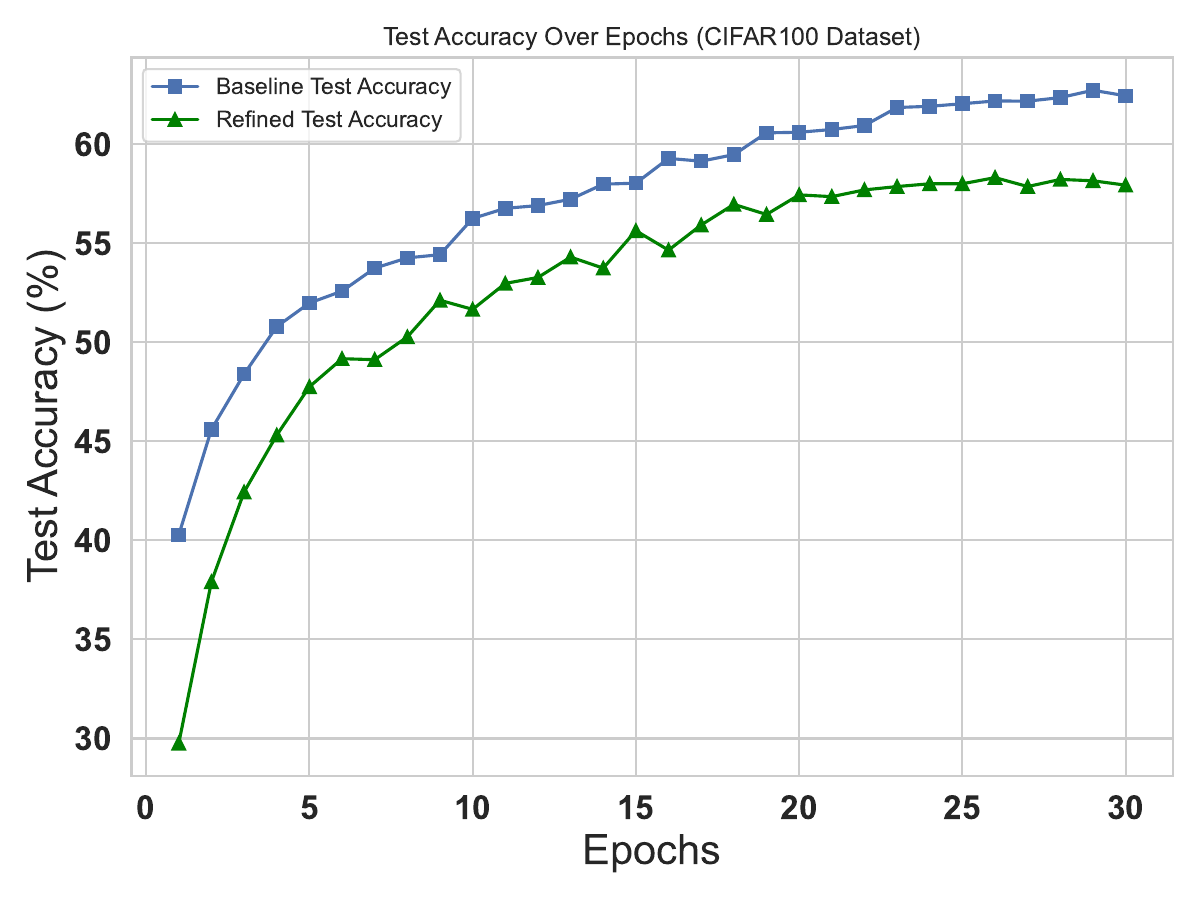}
    \caption{Test Accuracy Over Epochs for CIFAR100 Dataset. This plot shows the test accuracy for both the baseline and refined models over training epochs.}
    \label{fig:test_accuracy100}
\end{figure}

\begin{figure}[htbp]
    \centering
    \includegraphics[width=\linewidth]{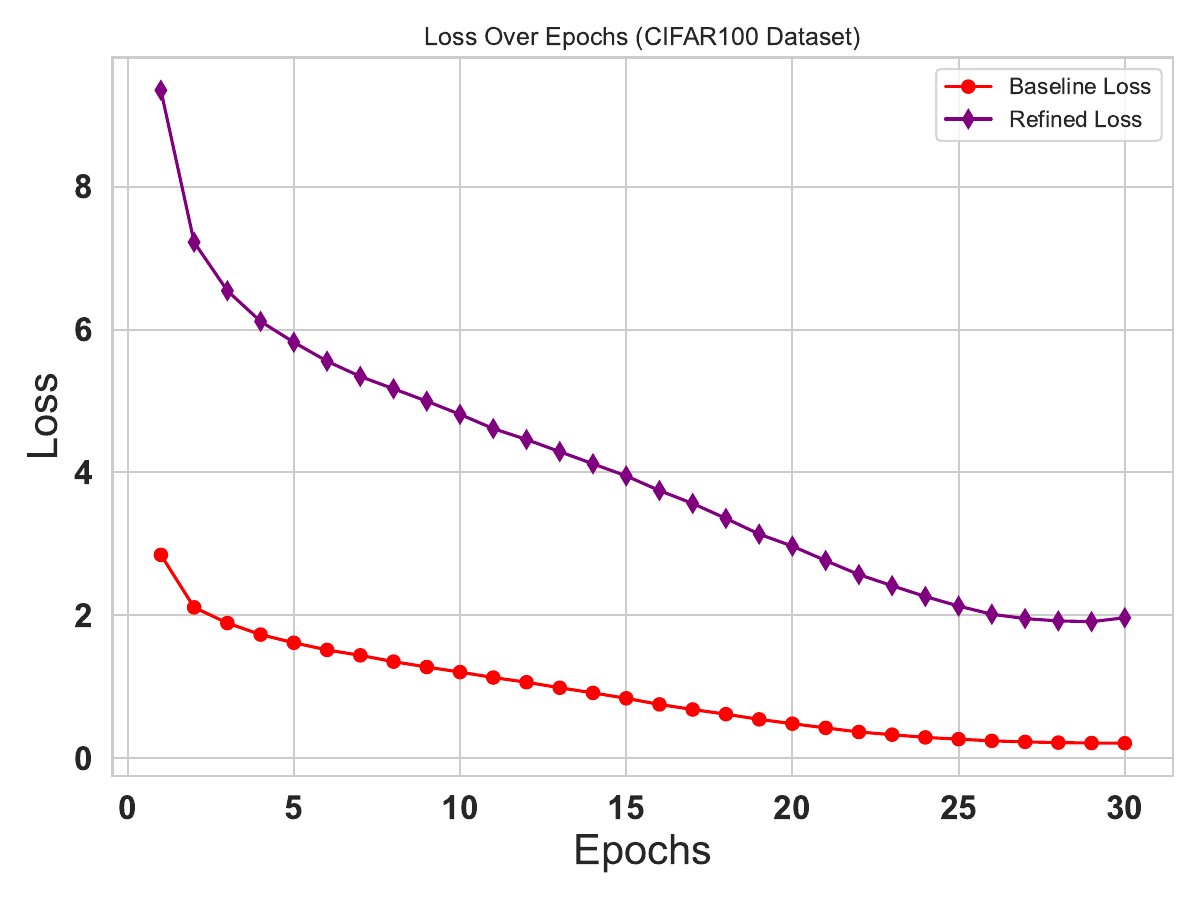}
    \caption{Loss Over Epochs for CIFAR100 Dataset. This plot illustrates the training loss for both the baseline and refined models over training epochs.}
    \label{fig:loss}
\end{figure}


\begin{figure}[htbp]
    \centering
    \includegraphics[width=\linewidth]{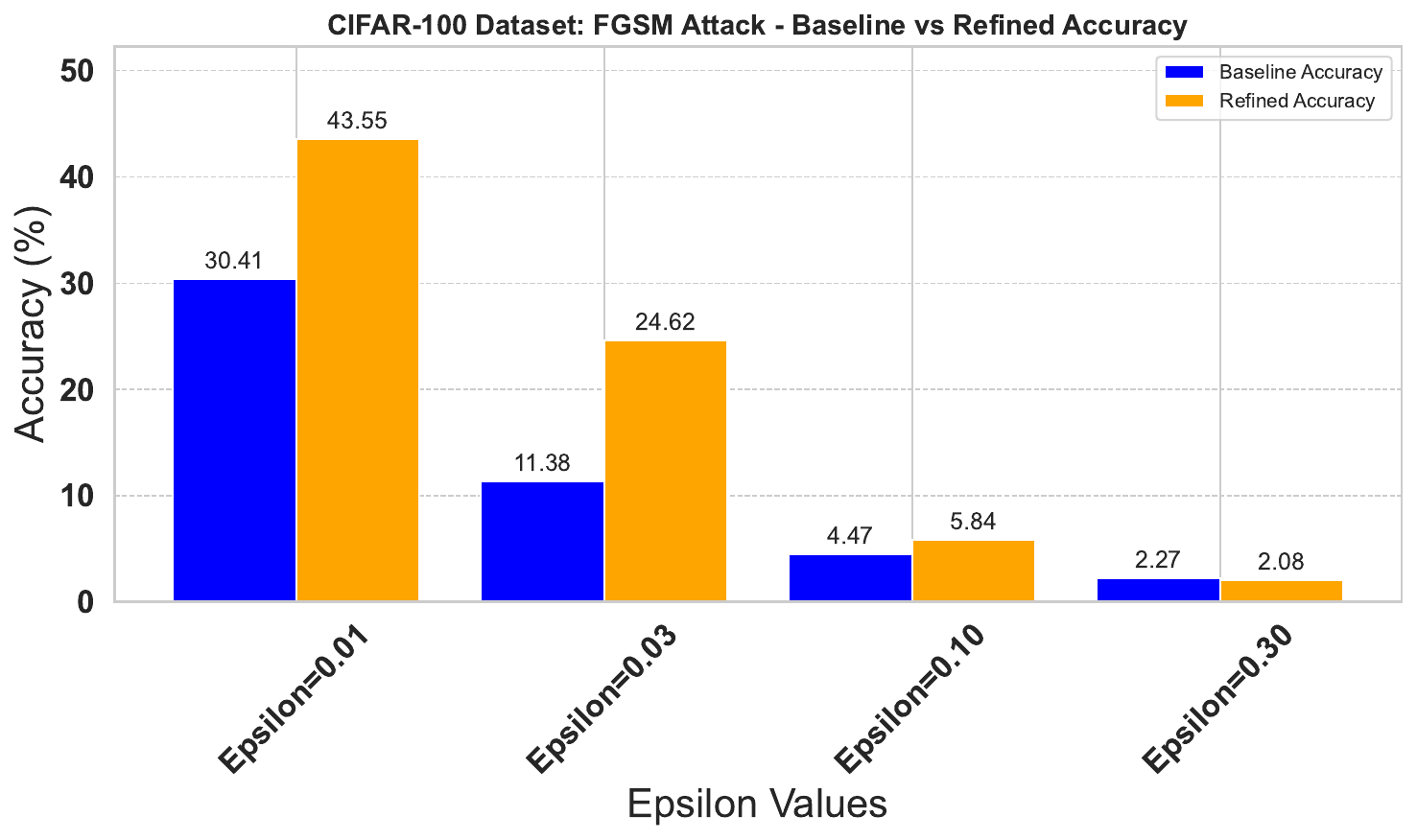}
    \caption{Comparison of Baseline and Refined Accuracy under FGSM Attack on the CIFAR-100 Dataset. The x-axis represents the epsilon values, and the y-axis shows the corresponding accuracies.}
    \label{fig:fgsm_plot100}
\end{figure}

\begin{figure}[htbp]
    \centering
    \includegraphics[width=\linewidth]{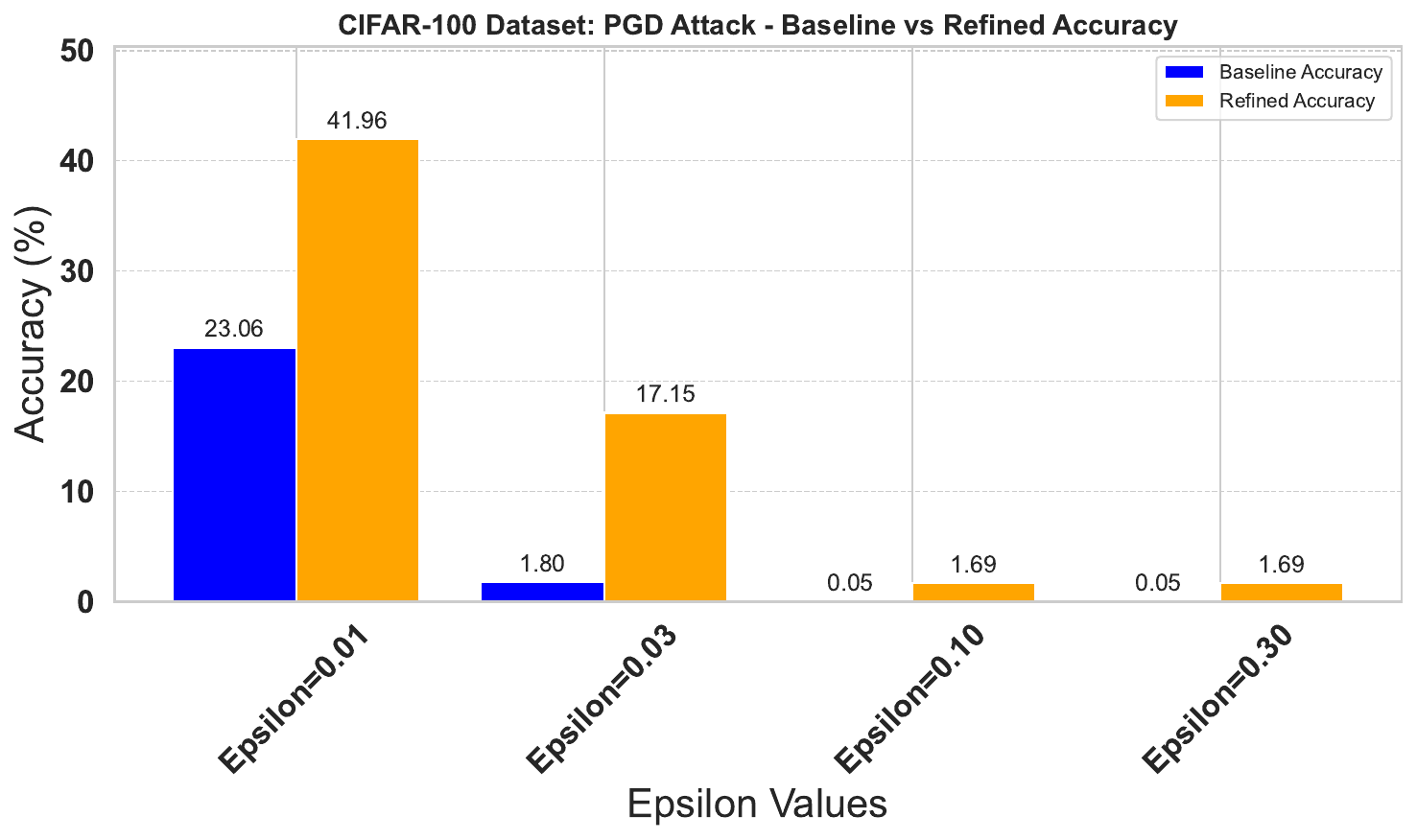}
    \caption{Comparison of Baseline and Refined Accuracy under PGD Attack on the CIFAR-100 Dataset. The x-axis represents the epsilon values, and the y-axis shows the corresponding accuracies.}
    \label{fig:pgd_plot100}
\end{figure}

We extend our experiments to the CIFAR-100 dataset to evaluate the performance of the LIME-guided refined model and the baseline model on a more challenging classification task. The CIFAR-100 dataset contains 100 classes, each with 500 training and 100 testing images. This increases the complexity of the learning process and provides a robust assessment of the models’ accuracy, loss convergence, and adversarial robustness.

\subsubsection{Test Accuracy Comparison}

Figure~\ref{fig:test_accuracy100} shows the test accuracy of the baseline and refined models over 30 training epochs. The baseline model achieves higher clean test accuracy, converging at approximately \textbf{61\%}. In contrast, the refined model stabilizes at \textbf{55\%}, exhibiting slower convergence and a reduced final accuracy compared to the baseline. The difference in accuracy is due to the additional constraints introduced during the refinement process, such as sensitivity regularization and adversarial training. These constraints force the model to focus on more robust and meaningful features, trading off some clean accuracy for improved robustness. Despite this, the refined model demonstrates consistent performance throughout the epochs after stabilization.

\subsubsection{Robustness to FGSM Attacks}

Figure~\ref{fig:fgsm_plot100} highlights the adversarial robustness of the two models under FGSM attacks with varying perturbation strengths ($\epsilon$). At a low perturbation magnitude of $\epsilon = 0.01$, the refined model achieves \textbf{43.55\%} adversarial accuracy, significantly outperforming the baseline model, which achieves only \textbf{30.41\%}. As the perturbation magnitude increases to $\epsilon = 0.03$, the refined model maintains an adversarial accuracy of \textbf{24.62\%}, while the baseline accuracy drops sharply to \textbf{11.38\%}. For larger perturbations, such as $\epsilon = 0.1$ and $\epsilon = 0.3$, both models experience a significant performance drop. However, the refined model retains a slight advantage over the baseline, demonstrating its ability to maintain robustness even under higher perturbations.

These results clearly demonstrate that the refined model is more resistant to FGSM attacks, particularly for small to moderate perturbation magnitudes. This improvement is a direct result of the iterative refinement process, which reduces the model’s reliance on vulnerable features.

\subsubsection{Robustness to PGD Attacks}

Figure~\ref{fig:pgd_plot100} presents the results of the models under PGD attacks, a stronger iterative adversarial method. At $\epsilon = 0.01$, the refined model achieves \textbf{41.41\%} accuracy, whereas the baseline model lags behind at \textbf{27.61\%}. As the perturbation magnitude increases to $\epsilon = 0.03$, the baseline model’s performance deteriorates rapidly, while the refined model retains an accuracy of \textbf{19.25\%}. For higher perturbation strengths, such as $\epsilon = 0.1$ and $\epsilon = 0.3$, both models experience near-complete degradation in performance, highlighting the challenge of defending against strong iterative attacks. Nevertheless, the refined model consistently outperforms the baseline across all perturbation levels, underscoring its enhanced robustness.

\subsection{CIFAR-10C Dataset}

\begin{figure}[htbp]
    \centering
    \includegraphics[width=\linewidth]{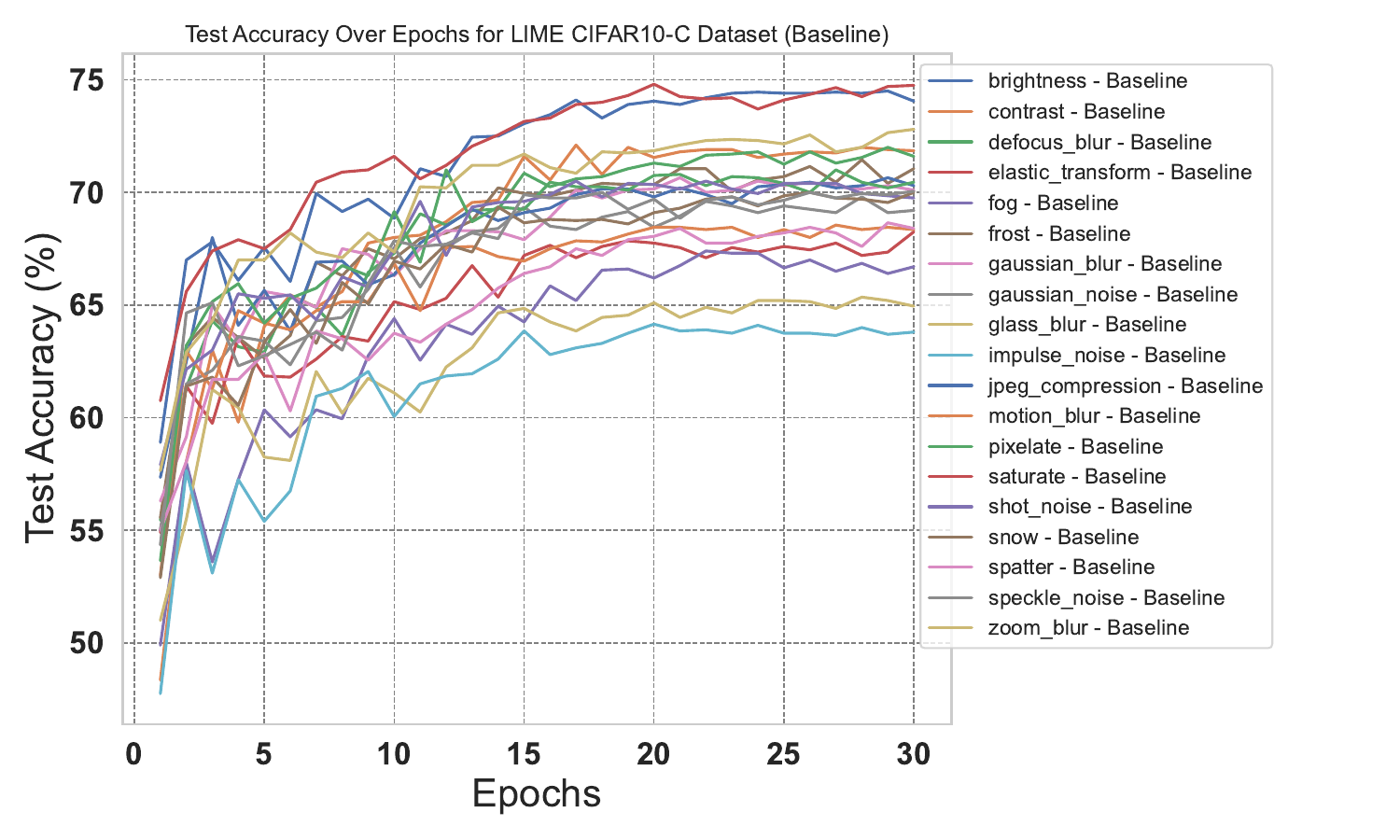}
    \caption{Test Accuracy Over Epochs for LIME CIFAR10-C Dataset (Baseline). Accuracy trends for various corruption types under the baseline approach are depicted.}
    \label{fig:baseline_accuracy10c}
\end{figure}


\begin{figure}[htbp]
    \centering
    \includegraphics[width=\linewidth]{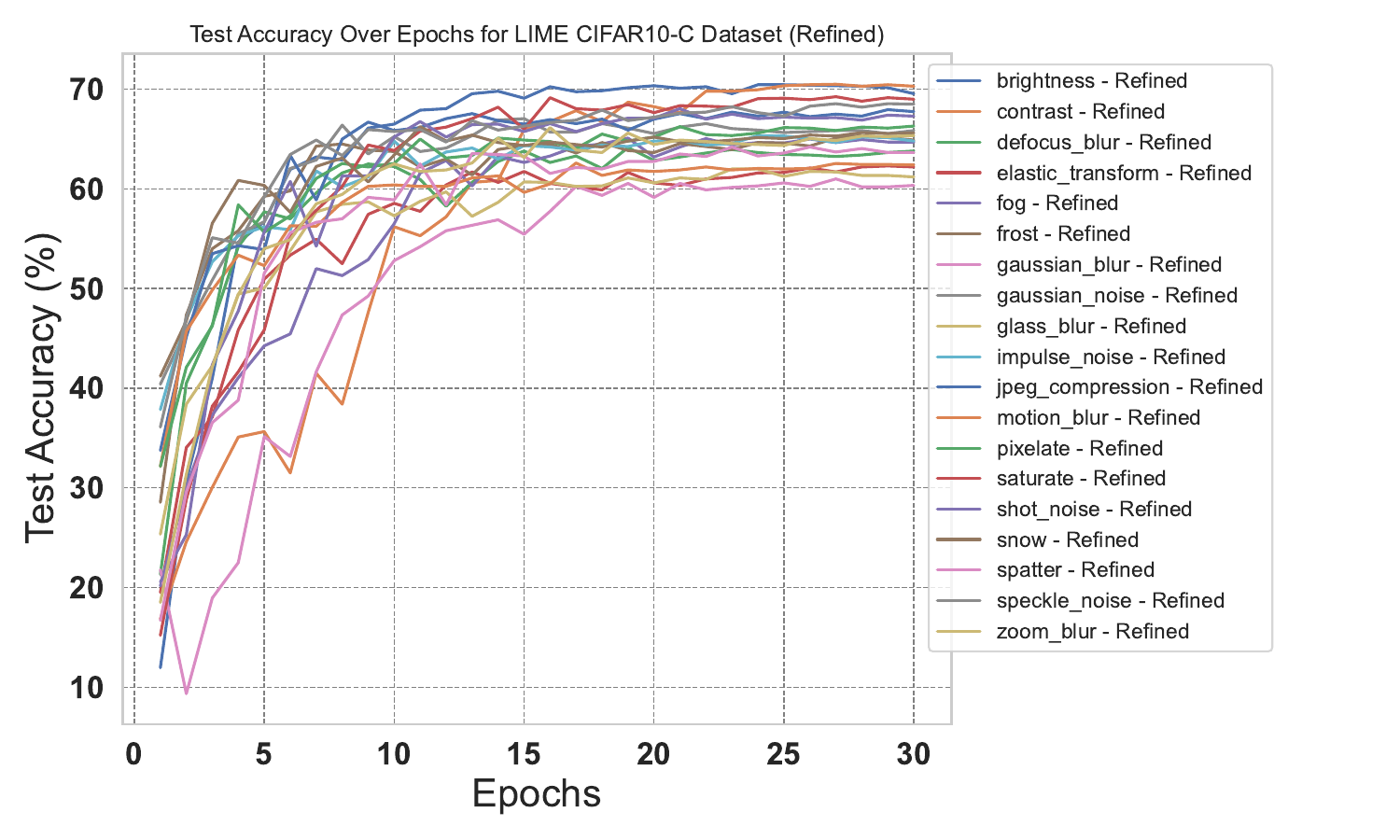}
    \caption{Test Accuracy Over Epochs for LIME CIFAR10-C Dataset (Refined). Accuracy trends for various corruption types under the refined approach are presented.}
    \label{fig:refined_accuracy10c}
\end{figure}





\begin{figure}[htbp]
    \centering
    \includegraphics[width=\linewidth]{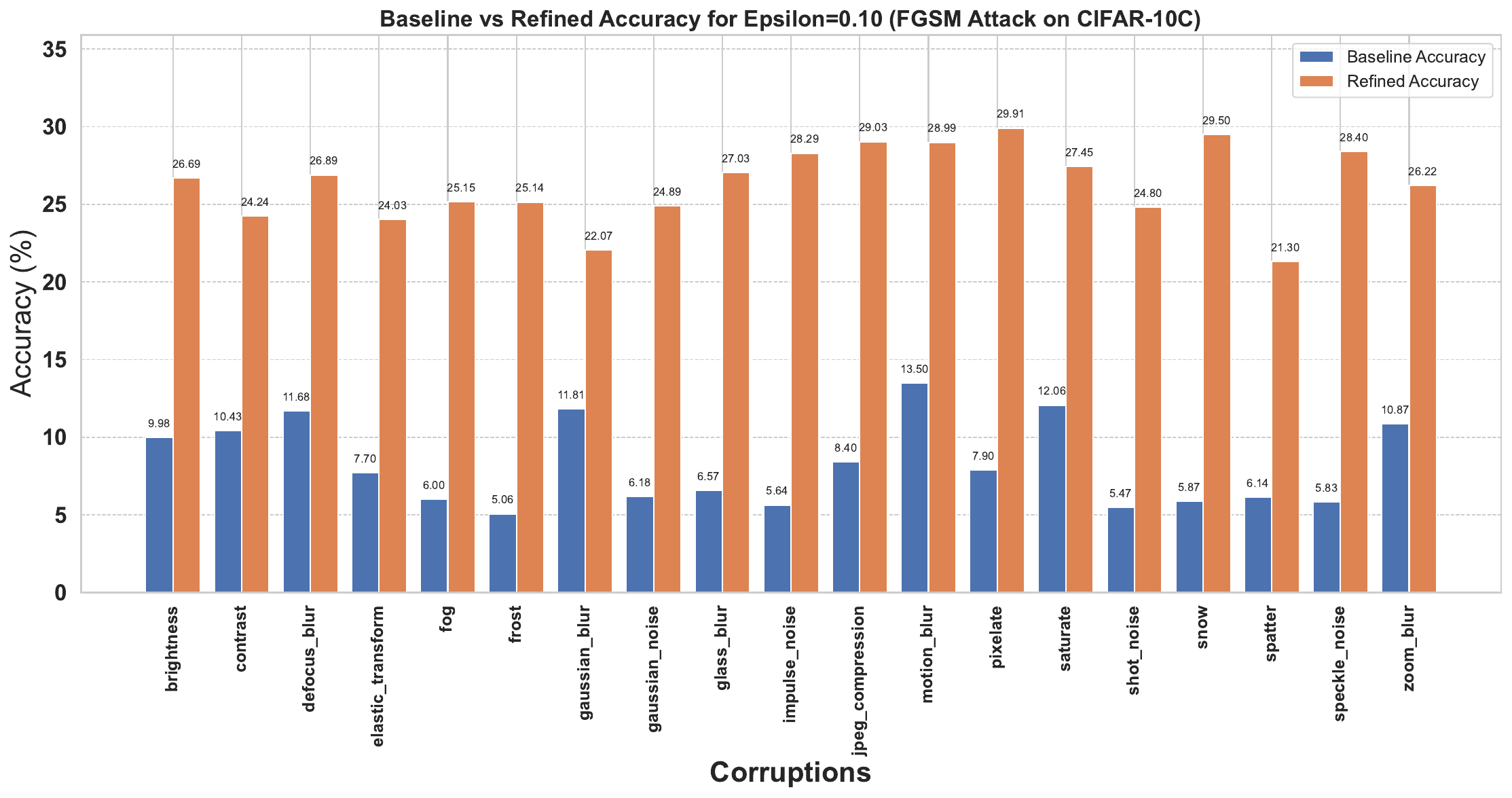}
    \caption{Baseline vs Refined Accuracy for Epsilon=0.10 (FGSM Attack) of LIME CIFAR10-C Dataset. The figure depicts reduced accuracy for higher epsilon values.}
    \label{fig:fgsm_eps_0.10}
\end{figure}

\begin{figure}[htbp]
    \centering
    \includegraphics[width=\linewidth]{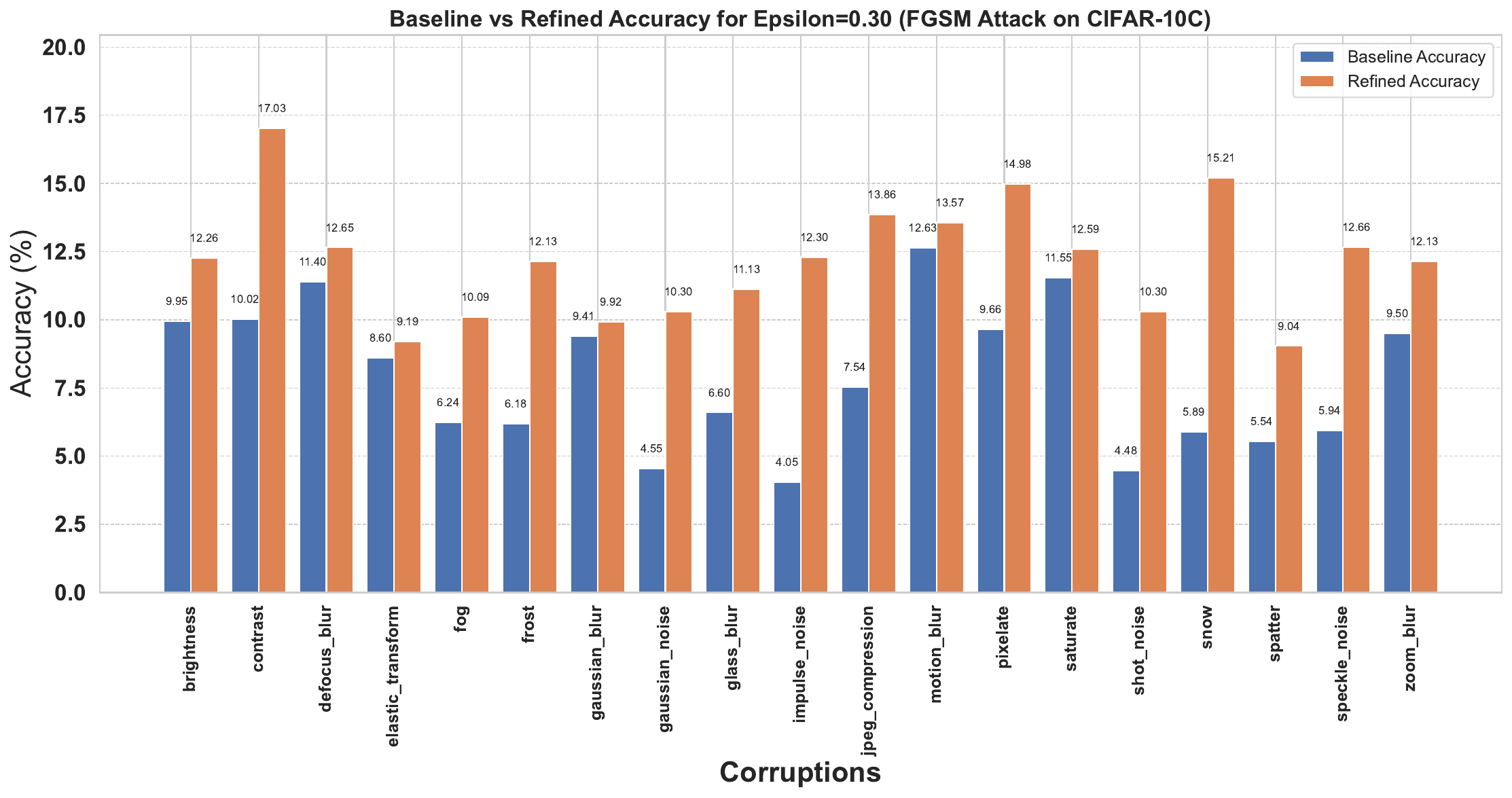}
    \caption{Baseline vs Refined Accuracy for Epsilon=0.30 (FGSM Attack) of LIME CIFAR10-C Dataset. The results indicate significant degradation in performance.}
    \label{fig:10cfgsm_eps_0.30}
\end{figure}

\begin{figure}[htbp]
    \centering
    \includegraphics[width=\linewidth]{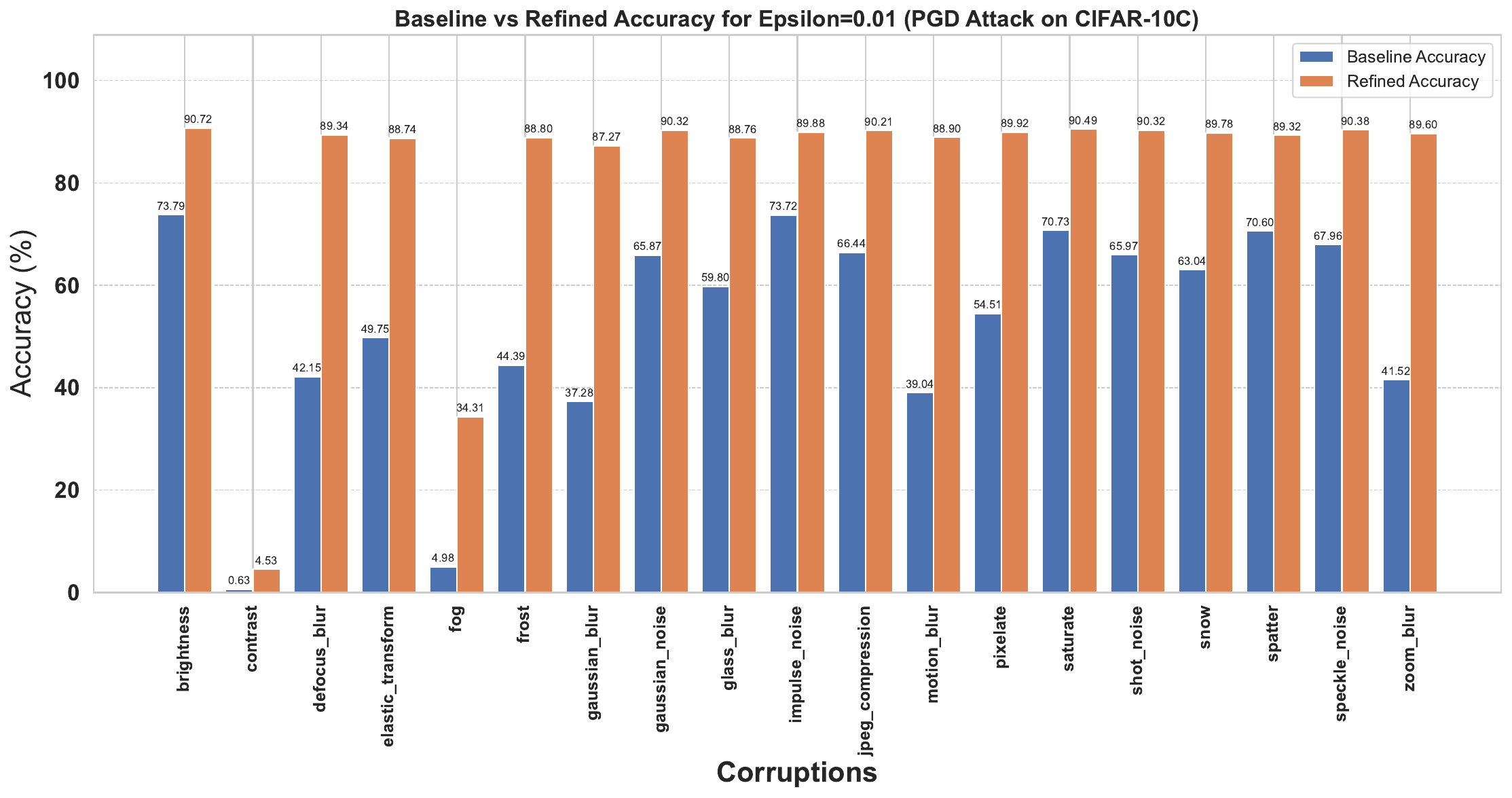}
    \caption{Baseline vs Refined Accuracy for Epsilon=0.01 (PGD Attack) of LIME CIFAR10-C Dataset. The plot compares accuracy for baseline and refined models.}
    \label{fig:10cpgd_eps_0.01}
\end{figure}

\begin{figure}[htbp]
    \centering
    \includegraphics[width=\linewidth]{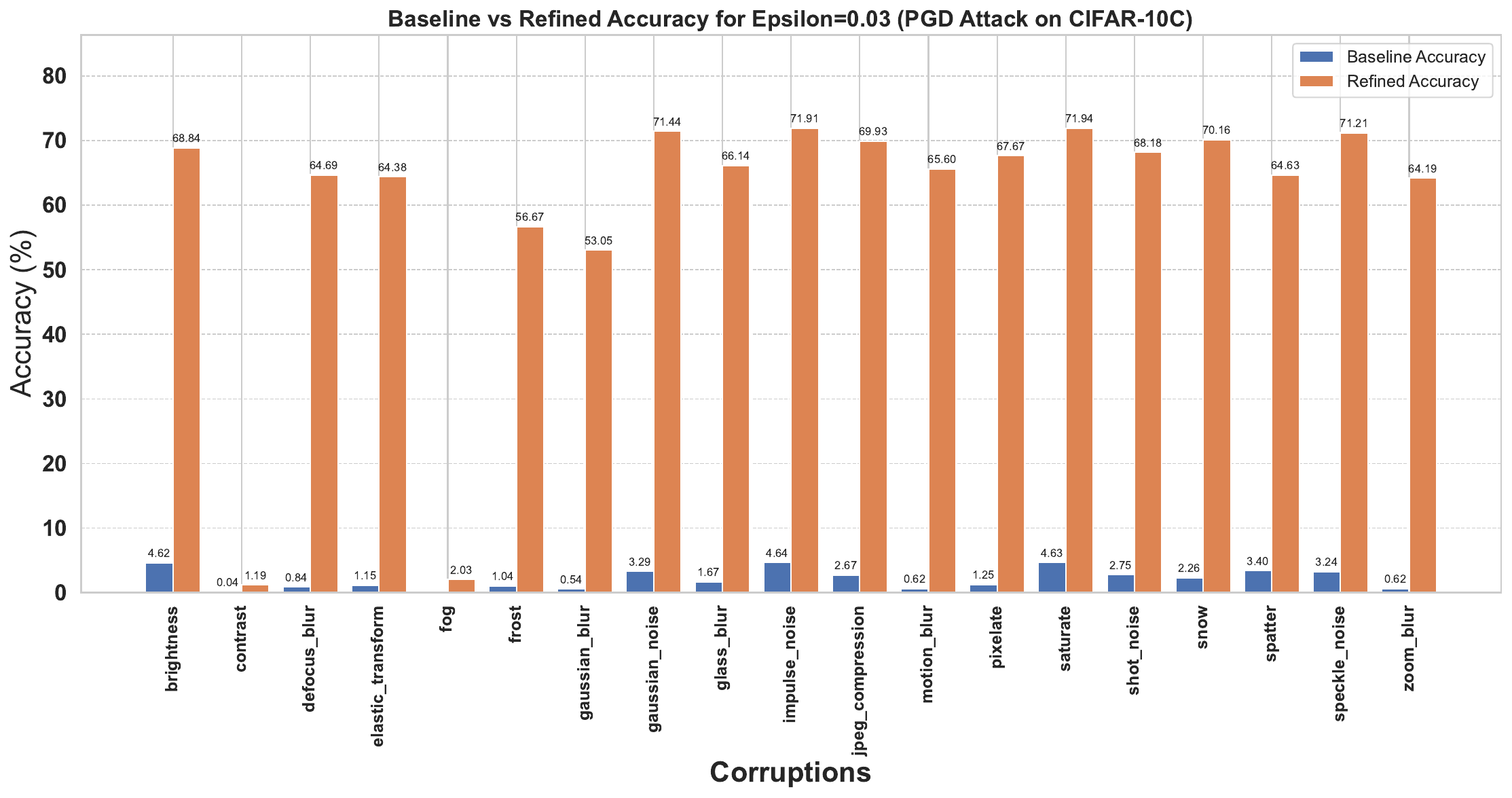}
    \caption{Baseline vs Refined Accuracy for Epsilon=0.03 (PGD Attack) of LIME CIFAR10-C Dataset. The figure shows accuracy trends across corruptions.}
    \label{fig:10cpgd_eps_0.03}
\end{figure}

To further validate the robustness of our LIME-guided refined model, we evaluate its performance on the CIFAR-10C dataset, a benchmark designed to assess model robustness under common corruptions. CIFAR-10C introduces 19 corruption types, such as noise, blur, weather distortions, and digital artifacts, across varying levels of severity. We compare the \textbf{baseline model} and the \textbf{refined model} in terms of clean accuracy, corruption robustness, and adversarial accuracy under FGSM and PGD attacks.

\subsubsection{Test Accuracy Under Common Corruptions}

Figure~\ref{fig:baseline_accuracy10c} and Figure~\ref{fig:refined_accuracy10c} illustrate the test accuracy of the baseline and refined models across the 19 corruption types over 30 training epochs. The baseline model achieves higher accuracy for some corruptions, peaking around \textbf{75\%} for specific cases. However, the refined model demonstrates more stable performance across the majority of corruption types, converging slightly below the baseline at around \textbf{70\%} accuracy.

The refined model's consistency stems from its reliance on robust, interpretable features rather than spurious patterns, which are often disrupted by corruptions. In contrast, the baseline model's higher variance across corruptions indicates its overfitting to less stable features, which compromises its resilience.

\subsubsection{Robustness to FGSM Attacks}

Figures~\ref{fig:fgsm_eps_0.10} and \ref{fig:10cfgsm_eps_0.30} compare the adversarial accuracy of the baseline and refined models under FGSM attacks at perturbation magnitudes $\epsilon = 0.1$ and $\epsilon = 0.3$.

At $\epsilon = 0.1$, the refined model outperforms the baseline model across all corruption types. For example, the refined model achieves \textbf{26.69\%} accuracy for brightness and \textbf{29.91\%} for impulse noise, compared to the baseline's \textbf{9.98\%} and \textbf{8.40\%}, respectively. The baseline model’s performance deteriorates sharply under perturbations, while the refined model retains a significant advantage, demonstrating its enhanced robustness.

At a larger perturbation magnitude of $\epsilon = 0.3$, the refined model continues to outperform the baseline across all corruptions. While both models exhibit a drop in performance, the refined model maintains higher accuracy, with results such as \textbf{17.03\%} for contrast and \textbf{15.21\%} for pixelate. In comparison, the baseline model’s accuracy remains below \textbf{10\%} for most corruption types.

\subsubsection{Robustness to PGD Attacks}

Figures~\ref{fig:10cpgd_eps_0.01} and \ref{fig:10cpgd_eps_0.03} show the models' performance under PGD attacks, which represent a stronger iterative adversarial method.

At a low perturbation magnitude of $\epsilon = 0.01$, the refined model achieves significantly higher accuracy than the baseline across all corruption types. For example, the refined model achieves \textbf{90.72\%} accuracy for brightness and \textbf{89.32\%} for speckle noise, compared to the baseline’s \textbf{73.79\%} and \textbf{41.52\%}, respectively. The substantial improvement highlights the refined model's ability to resist adversarial perturbations while maintaining robustness under corrupted inputs.

For $\epsilon = 0.03$, the refined model maintains superior performance, achieving consistent accuracy between \textbf{64\%} and \textbf{71\%} across all corruption types. In contrast, the baseline model’s accuracy deteriorates to near-zero values for most corruptions, indicating its inability to withstand adversarial attacks combined with input corruptions.

\section{Conclusion}

This paper presented a novel LIME-guided refinement framework to address critical vulnerabilities in deep learning models, such as susceptibility to adversarial attacks, reliance on spurious correlations, and limited interpretability. By leveraging LIME to identify and mitigate the influence of irrelevant features, the proposed approach systematically enhances both robustness and transparency. Empirical evaluations on CIFAR-10, CIFAR-100, and CIFAR-10C datasets demonstrated that LIME-guided models outperform baseline models under adversarial scenarios and input corruptions, showcasing significant improvements in adversarial resistance and generalization to out-of-distribution data. Future work will explore its application to other architectures and domains, as well as its integration with advanced defense techniques to further enhance model reliability and fairness.

\end{document}